\def\year{2019}\relax
\documentclass[letterpaper]{article} 
\usepackage{aaai19}  
\usepackage{times}  
\usepackage{helvet}  
\usepackage{courier}  
\usepackage{url}  
\usepackage{graphicx}  
\usepackage{amsmath}
\usepackage{amsfonts}
\usepackage{amssymb}
\usepackage{amsthm}
\usepackage{bbm}
\usepackage{mathtools}
\usepackage{blkarray}
\newtheorem{theorem}{Theorem}

\newtheorem*{corollary*}{Proof Sketch}
\newtheorem*{assumption*}{Assumption}
\newtheorem{definition}{Definition}

\usepackage[linesnumbered, algoruled]{algorithm2e}
\usepackage{subcaption}
\usepackage{booktabs}
\usepackage{subfiles}
\usepackage{color}
\begin{document}
%
\title{Less but Better: \\ Generalization Enhancement of Ordinal Embedding via Distributional Margin}
\author
{
		Ke Ma\textsuperscript{1,2}, Qianqian Xu\textsuperscript{3}, Zhiyong Yang\textsuperscript{1,2}, Xiaochun Cao\textsuperscript{1}\thanks{The corresponding authors.}\\
		\textsuperscript{1} State Key Laboratory of Information Security, Institute of Information Engineering, Chinese Academy of Sciences\\
		\textsuperscript{2} School of Cyber Security, University of Chinese Academy of Sciences\\
		\textsuperscript{3} Key Laboratory of Intelligent Information Processing, Institute of Computing Technology, Chinese Academy of Sciences\\
		\{make, yangzhiyong, caoxiaochun\}@iie.ac.cn, xuqianqian@ict.ac.cn\\
}
\maketitle
\begin{abstract}
	In the absence of prior knowledge, ordinal embedding methods obtain new representation for items in a low-dimensional Euclidean space via a set of quadruple-wise comparisons. These ordinal comparisons often come from human annotators, and sufficient comparisons induce the success of classical approaches. However, collecting a large number of labeled data is known as a hard task, and most of the existing work pay little attention to the generalization ability with insufficient samples. Meanwhile, recent progress in large margin theory discloses that rather than just maximizing the minimum margin, both the margin mean and variance, which characterize the margin distribution, are more crucial to the overall generalization performance. To address the issue of insufficient training samples, we propose a margin distribution learning paradigm for ordinal embedding,  entitled Distributional Margin based Ordinal Embedding (\textit{DMOE}). Precisely, we first define the margin for ordinal embedding problem. Secondly, we formulate a concise objective function which avoids maximizing margin mean and minimizing margin variance directly but exhibits the similar effect. Moreover, an Augmented Lagrange Multiplier based algorithm is customized to seek the optimal solution of \textit{DMOE} effectively. Experimental studies on both simulated and real-world datasets are provided to show the effectiveness of the proposed algorithm.
\end{abstract}

\noindent The problem of analyzing a set of $n$ objects given \textit{similarity} information is an inherent part in a broad variety of tasks in artificial intelligence \cite{Heikinheimo2013TheCA,503}, machine learning \cite{jamieson2011low,53e99af7b7602d97023851bf,arias-castro2017,DBLP:conf/nips/KleindessnerL17}, information retrieval \cite{DBLP:conf/aaai/LiuJWL16}, data mining \cite{DBLP:conf/sdm/LeL16} and computer vision \cite{wilberKKB2015concept}. Many algorithms are based on the assumption that `similar' inputs should generate `close' outputs. In a numerical setting of embedding, a similarity function (or, equivalently, a dissimilarity function) quantifies how `similar' objects are to others. The required input is the distance or similarity matrix of items. We calculate a set of embedded points which aims to preserve such similarities as well as possible. However, in recent years a whole new branch of the literature has emerged, which is the comparison-based embedding. Instead of evaluating similarity directly, we collect the similarity comparisons as follows: 

\emph{``Is the similarity between object $i$ and $j$ larger than the similarity between $l$ and $k$?''}

The corresponding problem is ordinal embedding. These two types of supervision information, numerical similarities and relative comparisons,  are all generated by human beings. Nevertheless, the latter one provides similarity estimates on a relative scale instead of the absolute scale. The comparison-based setting is a special case of the observation that humans are better at comparing two stimuli than at identifying a single one \cite{stewart2005absolute}. Consequently, the relative comparison is a more reliable form for incorporating human knowledge with artificial intelligence tasks.

The ordinal embedding problem was firstly studied by \cite{Shepard1962a,Shepard1962b,Kruskal1964a,Kruskal1964b} in the psychometric society. In recent years, it has drawn a lot of attention \cite{DBLP:conf/nips/SchultzJ03,agarwal2007generalized,tamuz2011adaptiive,vandermaaten2012stochastic,Terada2014LocalOE,amid2015multiview,doi:10.1137/1.9781611974010.31,DBLP:conf/nips/JainJN16,DBLP:conf/aaai/MaZXXCLY18}. One class of these typical methods is margin-based ordinal embedding which solves the problem under the classification framework. The well-known Generalized Non-Metric Multidimensional Scaling (\textit{GNMDS}) \cite{agarwal2007generalized} aims at finding a Gram matrix $\boldsymbol{G}$ such that the pairwise distances of embedded points satisfy the partial order constraints. Stochastic Triplet Embedding (\textit{STE/TSTE}) \cite{vandermaaten2012stochastic} is proposed to jointly penalize the violated constraints and reward the satisfied constraints via logistic loss. Multi-view Triplet Embedding (\textit{MTE}) \cite{amid2015multiview} decomposes the \textit{STE} objective function as different components and re-weights them for a better explanation. The other class of ordinal embedding methods uses the nearest neighbor graphs to model the similarity comparisons. Structure Preserving Embedding (\textit{SPE}) \cite{Shaw:2009:SPE:1553374.1553494} and Local Ordinal Embedding (\textit{LOE}) \cite{Terada2014LocalOE} embed unweighted nearest neighbor graphs to Euclidean spaces with convex and non-convex objective functions. The nearest neighbor adjacency matrix can be transformed into ordinal constraints, but it is not a standard equipment in comparison-based scenarios. With this limitation, \textit{SPE} and \textit{LOE} are not suitable for ordinal embedding via quadruplets or triple comparisons.

A common issue of the existed ordinal embedding methods is the dependence of large samples of similarity comparisons. \cite{53e99af7b7602d97023851bf,arias-castro2017} show the consistency of ordinal embedding problem. When the number of the objects $n$ tends to infinity, the set of embedded points always converges to the set of original points, up to similarity transformations; the rate of convergence depends on the Hausdorff distance between the ground-truth points. Later \cite{DBLP:conf/nips/JainJN16} show a finite sampling result of consistency. Learning an embedding which predicts nearly as well as the true embedding needs $\boldsymbol{\Theta}(pn\log n)$ samples, where $p$ is the embedding dimension. There is a strong condition that the triple-wise comparisons are generated from the classical Bradley-Terry-Luce (\textit{BTL}) model \cite{10.2307/2334029,luce_individual_1959} and this assumption could not be verified in the actual applications. The theoretical results suggest that only the adequateness of similarity comparisons can promise the prediction result. However, the cost of eliciting relative similarity comparisons from human beings would be prohibitive. The amenable applications for collecting the relative similarity comparisons, e.g., crowdsourcing and human computation, need passively waiting for participants and stimulate them with money to get the desired information. Without prior knowledge, the relative comparisons always involve all objects, and the number of possible comparisons could be $\boldsymbol{\Theta}(n^4)$. The spending of data collection presents ordinal embedding methods with a dilemma: the insufficient samples would limit the potential performance; the adequate samples with prospective results would be cumbersome. Unfortunately, most of the traditional methods ignore that the generalization is the main concern in ordinal embedding task with insufficient samples. 

In this paper, we propose a new method, named Distributional Margin based Ordinal Embedding (\textit{DMOE}), which tries to achieve strong generalization performance by optimizing the margin distribution in ordinal embedding problem. Inspired by the recent results in classification \cite{DBLP:conf/kdd/ZhangZ14,DBLP:conf/bibm/YangLZ16}, we define the margin of ordinal embedding and characterize the margin distribution by the first- and second-order statistics, and try to maximize the margin mean and minimize the margin variance simultaneously. For optimization, we propose an alternating direction method of multipliers (\textit{ADMM}) for \textit{DMOE} with semi-definite and low-rank constraints. Comprehensive experiments on the synthetic and real-world datasets show the superiority of our method to other ordinal embedding algorithms, verifying that the margin distribution is more crucial for generalization than minimum margin.


\section{Problem Definition}

Throughout the paper, scalars, vectors, matrices and sets are denoted as lowercase letters ($x$), bold lower case letters ($\boldsymbol{x}$), bold capital letters ($\boldsymbol{X}$) and calligraphy uppercase letters ($\mathcal{X}$). $x_{ij}$ denotes the $(i, j)$ entry of $\boldsymbol{X}$. $[n]$ is the set of $\{1,\dots,n\}$. $\mathbb{E}(\cdot)$ represents the expectation. 

Suppose $\mathcal{O} = \{\boldsymbol{o}_1,\dots,\boldsymbol{o}_n\}$ is a set of $n$ objects, we assume that a certain but unknown similarity function $\zeta:\mathcal{O}\times\mathcal{O}\rightarrow\mathbb{R}^{+}$ assigns similarity value $\zeta_{ij}$ for a pair of objects $(\boldsymbol{o}_i, \boldsymbol{o}_j)$. With similarity function $\zeta$, a quadruplet $q=(i,j,l,k)$ defines the corresponding ordinal constraint, and these constraints lead to the ordinal embedding problem.

\begin{definition}[Ordinal Constraints]
	Given a set of quadruplets
	\begin{equation}
		\begin{aligned}
			& \mathcal{Q}=\{q\ \ \ |& &q = (i,j,l,k),(i,j)\neq(l,k),\\
			& && i\neq j,\ l\neq k,\ i,j,l,j\in[n]\ \}
		\end{aligned}
	\end{equation}	
	which is a subset of $[n]^4$, the ordinal constraints $\mathcal{Y}_{\mathcal{Q}}=\{y_q|q\in\mathcal{Q}\}\subset\{-1, +1\}^{|\mathcal{Q}|}$, implies the similarity partial order of object pairs in $\mathcal{O}$ as 
	\begin{equation}
		y_q=
		\left\{
		\begin{matrix}
		+1, & \text{if}\ \ \zeta_{ij}\ >\ \zeta_{lk},\\ 
		-1, & \text{if}\ \ \zeta_{ij}\ <\ \zeta_{lk}.
		\end{matrix}
		\right.
	\end{equation}
\end{definition}

Our goal here is to obtain a set of embedded points $\boldsymbol{X}$ which satisfy the ordinal constraints $\mathcal{Q}$. Without prior knowledge, embedding $\mathcal{O}$ into a Euclidean space $\mathbb{R}^p$ is the most common situation which assumes that the squared Euclidean distances among embedded points are inversely proportional to the unknown similarity values. Specifically, a large distance of two embedded points $d^2_{ij}=\|\boldsymbol{x}_i-\boldsymbol{x}_j\|^2_2$ means the corresponding objects $\boldsymbol{o}_i$ and $\boldsymbol{o}_j$ would have small similarity value $\zeta_{ij}$. This assumption connects the squared Euclidean distances of $\boldsymbol{X}$ and ordinal constraints $\mathcal{Q}$. We further give the formal definition of ordinal embedding.

\begin{definition}[Ordinal Embedding]
	Suppose $\mathcal{Q}$ is a collection of quadruplets which are drawn independently and uniformly at random and $\mathcal{Y}_{\mathcal{Q}}$ is the correspondence ordinal constraints of object set $\mathcal{O}$. Let 
	$
		\boldsymbol{X}=\{\boldsymbol{x}_1,\dots,\boldsymbol{x}_n\}\in\mathbb{R}^{p\times n}
	$
	is the desired embedding in the Euclidean space $\mathbb{R}^{p}$ where $p\ll n$ and
	$
		\boldsymbol{D} = \{d^2_{ij}\}\in\mathbb{R}^{n \times n}
	$
	is the squared Euclidean distance matrix of embedding $\boldsymbol{X}$. Ordinal embedding is the problem of obtaining $\boldsymbol{X}$ with ordinal constraints $\mathcal{Y}_{\mathcal{Q}}$ on $\boldsymbol{D}$ such that
	$$
		\text{sign}(y_q\cdot\Delta_q \boldsymbol{D})>0,\ \forall\ y_q\in\mathcal{Y}_{\mathcal{Q}},
	$$
	where 
	$$
		\Delta_q \boldsymbol{D} = d^2_{ij}-d^2_{lk} = \|\boldsymbol{x}_i-\boldsymbol{x}_j\|^2_2-\|\boldsymbol{x}_l-\boldsymbol{x}_k\|^2_2.
	$$
\end{definition}
Note that one cannot consistently estimate the underlying embedding $\boldsymbol{X}$ with only ordinal supervision and without direct observations. In the case when no direct measurements are available, say the metric information of $\mathcal{O}$ as the input, the underlying embedding $\boldsymbol{X}$ is only identifiable up to certain monotonic transformations, \textit{e.g.}, rotation, reflection, translation, and scaling. Therefore, the sign consistency is adopted as the goal of ordinal embedding. 

By the above definition, the margin of instance $(\boldsymbol{X}_q, y_q)$ can be naturally defined as
\begin{equation}
	\label{eq:margin_D}
	\gamma_q = y_q\cdot\Delta_q \boldsymbol{D},
\end{equation}
where $\boldsymbol{X}_q = \{\boldsymbol{x}_i, \boldsymbol{x}_j, \boldsymbol{x}_l, \boldsymbol{x}_k\},\ \forall\ q\in\mathcal{Q}$.

Despite the close relationship between $\boldsymbol{D}$ and $\boldsymbol{X}$, $\Delta_q \boldsymbol{D}$ is a nonlinear function of $\boldsymbol{X}$ and it always leads to a non-convex optimization problem. Here we introduce the Gram matrix of $\boldsymbol{X}$ and construct a margin function as a linear function of Gram matrix. Firstly, a map is established to connect the distance matrix $\boldsymbol{D}$ and the Gram matrix $\boldsymbol{G}=\boldsymbol{X}^\top\boldsymbol{X}$: 
\begin{equation*}
	\begin{aligned}
		& d_{ij} &=&\ \ g_{ii}-2g_{ij}+g_{jj}\\
		& \boldsymbol{D}&=&\ \ \textit{diag}(\boldsymbol{G})\cdot\boldsymbol{1}^\top-2\boldsymbol{G}+\boldsymbol{1}\cdot\textit{diag}(\boldsymbol{G})^\top,
	\end{aligned}
\end{equation*}
 where $\textit{diag}(\boldsymbol{G})$ is the column vector composed of the diagonal entries of $\boldsymbol{G}$ and $\boldsymbol{1}$ is the $n$-dimension vector with all entries being $1$. With a little abuse of $\Delta_q $, the margin of instance $(\boldsymbol{X}_q, y_q)$ can be written as
\begin{equation}
	\label{eq:margin_G}
	\begin{aligned}
		& \gamma_q &=&\ \ y_q\cdot\Delta_q \boldsymbol{D} = y_q(d_{ij}-d_{lk})\\
		& &=&\ \ y_q(g_{ii}-2g_{ij}+g_{jj}-g_{ll}+2g_{lk}-g_{kk})\\
		& &\coloneqq&\ \ y_q\cdot\Delta_q \boldsymbol{G},\ \ \ \ \forall\ q\in\mathcal{Q}.
	\end{aligned}
\end{equation}

By the definition of \eqref{eq:margin_G}, ordinal embedding can be formulated as the following convex optimization problem.
\begin{definition}[The Margin-based Ordinal Embedding]
	Let $l:\mathbb{R}^{+}\rightarrow\mathbb{R}$ be a loss function which satisfies
	\begin{equation}
		l(x):
		\left\{
		\begin{matrix}
			\ >\ 0,\ \ & \text{if }x<0,\\ 
			\ \leq\ 0,\ \ & \text{if }x>0.
		\end{matrix}
		\right.
	\end{equation}
	Given the ordinal constraints $\mathcal{Y}_{\mathcal{Q}}$, the ordinal embedding problem can be formulated as a semi-definite programming of Gram matrix $\boldsymbol{G}$:
	\begin{equation}
		\begin{aligned}
		\label{eq:sdp_oe}
			& &\underset{\boldsymbol{G}}{\min}&\ \ L(\boldsymbol{G}, \mathcal{Y}_{\mathcal{Q}})\\
			& &s.t.&\ \ \boldsymbol{G}\succeq 0,\ \ \textit{rank}(\boldsymbol{G})\leq p, 
		\end{aligned}
	\end{equation}
	where $L(\boldsymbol{G}, \mathcal{Y}_{\mathcal{Q}}) = \frac{1}{|\mathcal{Q}|}\underset{q\in\mathcal{Q}}{\sum}\ \ l(\gamma_q)$.
\end{definition}
We note that \eqref{eq:sdp_oe} is a semi-definite programming (\textit{SDP}) and $\boldsymbol{G}\succeq 0$ comes from the fact that $\boldsymbol{G}=\boldsymbol{X}^\top\boldsymbol{X}$ is a positive semi-definite matrix. Furthermore, the desired embedding dimension $p$ is a parameter of the ordinal embedding. It is well known that there exists a perfect embedding $\boldsymbol{X}$ estimated by any label set $\mathcal{Y}$ on the Euclidean distances in $\mathbb{R}^{n-2}$, even for the noisy constraints \cite{borg2003modern}. However, the low-dimensional setting where $p\ll n$ is the main task of this work. The smallest $p$ for noisy ordinal constraints $\mathcal{Y}_\mathcal{Q}$ is a future direction which worths pursuing. The choice of $p$ in the experiment section depends on the potential applications.

For example, the Generalized Non-metric Multidimensional Scaling (\textit{GNMDS}) follows the \textit{SVM} formulation to obtain the $\boldsymbol{G}$ by solving
\begin{equation}
	\label{opt:gnmds}
	\begin{aligned}
		& &\underset{\boldsymbol{G},\ \boldsymbol{\xi}}{\min}&\ \ \frac{1}{|\mathcal{Q}|}\sum_{q\in\mathcal{Q}}\xi_q\\
		& &s.t.&\ \ \gamma_q\geq\gamma_0-\xi_q,\ \xi_q\geq0,\ \forall\ q\in\mathcal{Q}, \\
		& & &\ \ \boldsymbol{G}\succeq 0,\ \ \text{rank}(\boldsymbol{G})\leq p, 
	\end{aligned}
\end{equation}
where $\gamma_0$ is a relaxed minimum margin and $\boldsymbol{\xi}=\{\xi_q\}_{q\in\mathcal{Q}}$ is the slack variable.

\section{Distributional Margin based Embedding}
The relaxed minimum margin $\gamma_0$ in GNMDS indeed characterizes the top minimum margins of all instance $\{\boldsymbol{X}_q, y_q\}_{q\in\mathcal{Q}}$. In margin theory of classification, it is known that maximizing the minimum margin of training examples is not sufficient to achieve fulfilling generalization performance \cite{DBLP:conf/icml/ReyzinS06}. The margin distribution of training examples, rather than the minimum margin, is more crucial to generalization performance in classification \cite{DBLP:journals/ai/GaoZ13a,DBLP:journals/corr/ZhangZ16a}. 
\subsection{Formulation}
The two most usual statistics for characterizing the margin distribution are the first- and second-order statistics, that is, the mean and the variance of margin. According to \eqref{eq:margin_G}, the margin mean of training samples $\{(\boldsymbol{X}_q, y_q)\}$ is
\begin{equation}
	\label{eq:margin_mean}
	\begin{aligned}
		& \bar{\gamma} &=&\ \ \frac{1}{|\mathcal{Q}|}\sum_{q\in\mathcal{Q}} \gamma_q\\
	\end{aligned}
\end{equation}
and the margin variance is 
\begin{equation}
	\label{eq:margin_variance}
	\hat{\gamma} = \frac{1}{|\mathcal{Q}|}\sum_{q\in\mathcal{Q}}(\gamma_q-\bar{\gamma})^2.
\end{equation}
Intuitively, we attempt to maximize the margin mean and minimize the margin variance simultaneously in ordinal embedding problem \eqref{eq:sdp_oe}.

First, there is a straightforward idea to achieve our goal as considering the margin mean \eqref{eq:margin_mean} and the margin variance \eqref{eq:margin_variance} in \eqref{eq:sdp_oe} explicitly. Although \eqref{eq:sdp_oe} can adopt different loss functions, we will focus on \textit{SVM} formulation \eqref{opt:gnmds} because the hinge loss is a natural form of margin. Considering the margin distribution, the optimization problem \eqref{opt:gnmds} can be formulated as 
\begin{equation}
	\label{opt:ldm}
	\begin{aligned}
		& &\underset{\boldsymbol{G},\ \boldsymbol{\xi}}{\min}&\ \ \frac{1}{|\mathcal{Q}|}\sum_{q\in\mathcal{Q}}\xi_q-\lambda_1\bar{\gamma}+\lambda_2\hat{\gamma}\\
		& &s.t.&\ \ \gamma_q\geq\gamma_0-\xi_q,\ \xi_q\geq0,\ \forall\ q\in\mathcal{Q}, \\
		& & &\ \ \boldsymbol{G}\succeq 0,\ \textit{rank}(\boldsymbol{G})\leq p,
	\end{aligned}
\end{equation}
where $\lambda_1$ and $\lambda_2$ are the trade-off parameters for balancing the impacts of $\bar{\gamma}$ and $\hat{\gamma}$. It is apparent that \textit{GNMDS} \eqref{opt:gnmds} is a degenerate case of \eqref{opt:ldm} when $\lambda_1$ and $\lambda_2$ equal to $0$. However, there exists an obvious drawback of \eqref{opt:ldm} with directly optimizing the margin distribution: tuning the parameters, $\lambda_1$ and $\lambda_2$, is an obstacles of solving \eqref{opt:ldm} efficiently. Therefore, a new lightweight formulation is proposed to optimize margin distribution implicitly. 

Recall that \textit{SVM} fixes the minimum margin as $1$ by scaling the margin with the norm of linear predictor. Following the similar way, we can scale the margin of $(\boldsymbol{X}_q, y_q)$ in ordinal embedding \eqref{eq:margin_G} and set the margin mean as a constant. This would not result in a sub-optimal solution because the ordinal constraints $\mathcal{Y}_\mathcal{Q}$ can only determine an embedding $\boldsymbol{X}$ up to the monotonic transformations. Without loss of generality, the mean of $\boldsymbol{\gamma}_{\mathcal{Q}}=\{\gamma_q|q\in\mathcal{Q}\}$ can be set as a constant and an equality constraint is conducted
\begin{equation}
	\label{eq:mean_constraint}
	\frac{1}{|\mathcal{Q}|}\underset{q\in\mathcal{Q}}{\sum}\gamma_q = \tilde{\gamma}_0.
\end{equation}
On the other hand, we want to minimize the variance of $\boldsymbol{\gamma}_{\mathcal{Q}}$. By \eqref{eq:mean_constraint}, the deviation of $\gamma_q$ to the margin mean $\tilde{\gamma}_0$ is {$|\gamma_q-\tilde{\gamma}_0|$}, and we force the deviation to be smaller than $\varepsilon_q\geq 0$ as
\begin{equation}
	\label{eq:variance_constraint}
	|\gamma_q-\tilde{\gamma}_0|\leq\varepsilon_q,\ \forall\ q\in\mathcal{Q}.
\end{equation}
Thus, minimizing $\varepsilon_q$ is equivalent to minimize the margin variance \eqref{eq:margin_variance}. Meanwhile, \eqref{eq:variance_constraint} implies the margin mean constraint \eqref{eq:mean_constraint}. 

\begin{figure}[thb!]
	\centering
	\begin{subfigure}[b]{0.48\columnwidth}
		\centering
		\includegraphics[width = \textwidth]{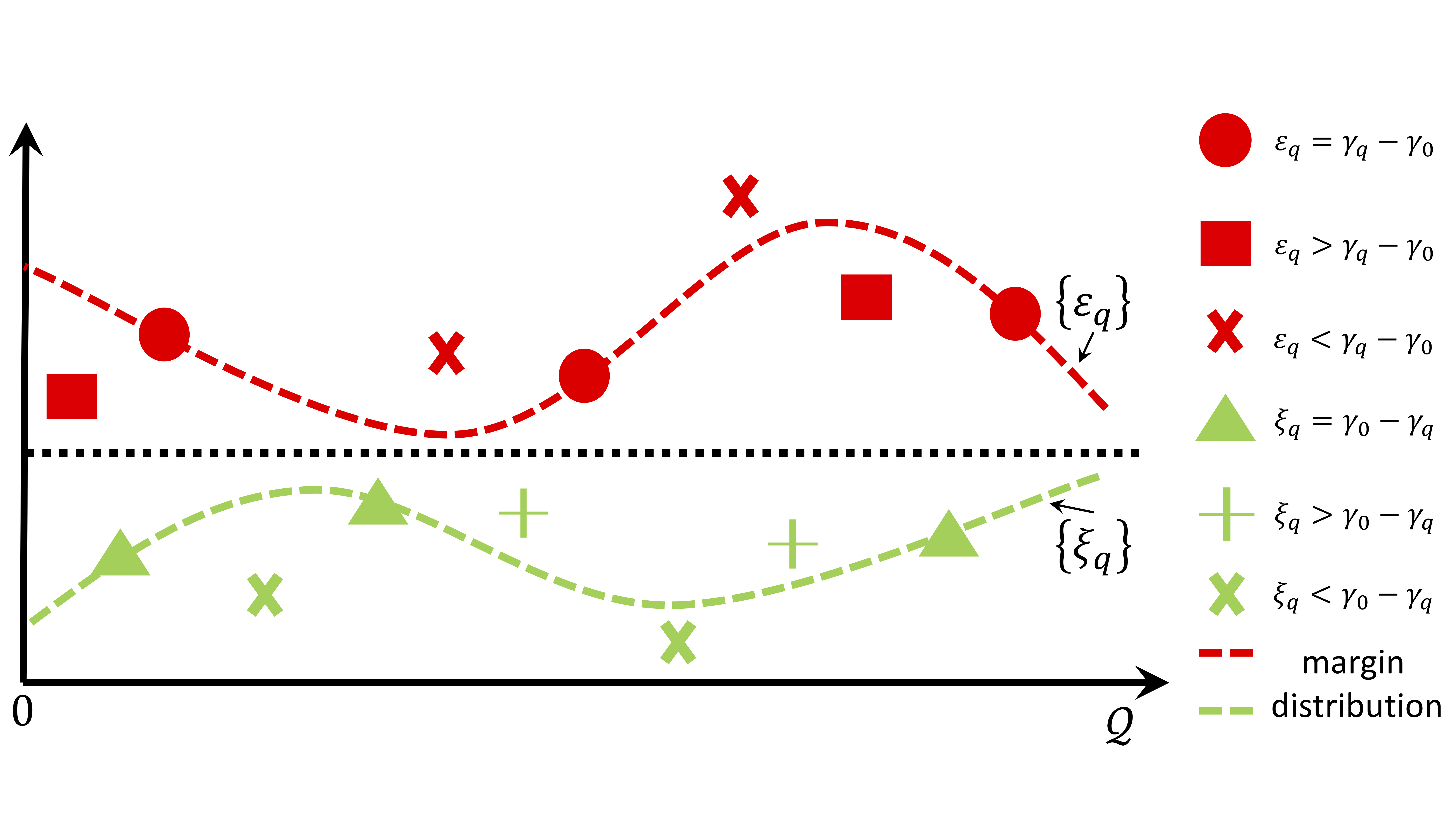}
		\caption{Margin Distribution}
		\label{fig:margin_distribution}
	\end{subfigure}
	\begin{subfigure}[b]{0.48\columnwidth}
		\centering
		\includegraphics[width = \textwidth]{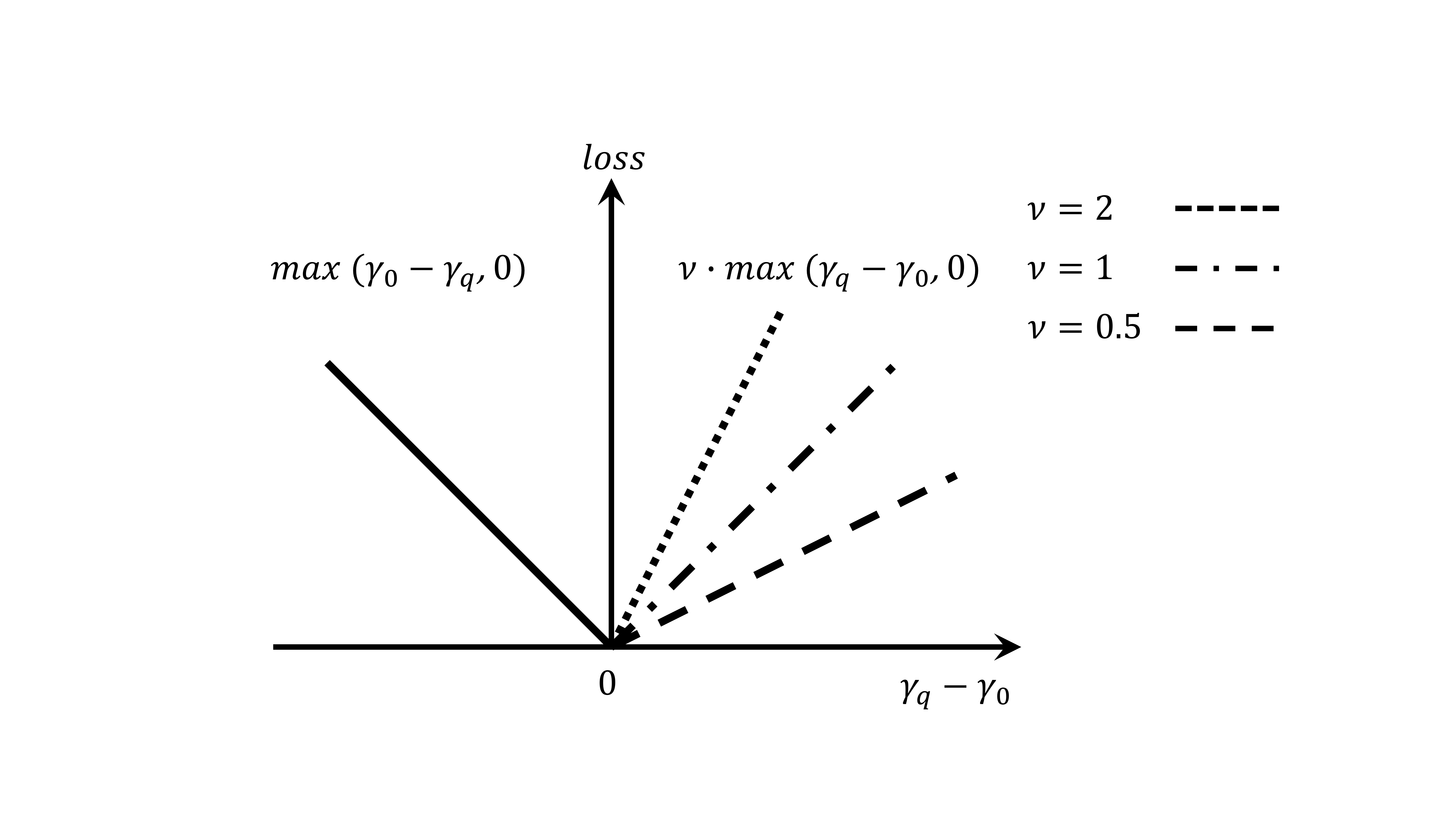}
		\caption{Loss Function}
		\label{fig:heterogenous_loss}
	\end{subfigure}
	\caption{(a) The Margin Distribution obtained by solving \eqref{opt:odm_1}. (b) The loss function \eqref{eq:bisides-hinge} with different $\nu$.}
	\label{fig:distribution_loss}
\end{figure}

Constraints like \eqref{eq:variance_constraint} in optimization problems always involve two inequality, $\forall\ q\in\mathcal{Q}$, 
\begin{subequations}
	\label{eq:margin_constraint}
	\begin{align}
		\gamma_q&\leq\tilde{\gamma}_0 +\varepsilon_q \label{eq:varian_seq_1},\\
		\gamma_q&\geq\tilde{\gamma}_0-\varepsilon_q \label{eq:varian_seq_2}.
	\end{align}
\end{subequations}
Note that the soft-margin constraint 
\begin{equation}
	\label{eq:soft_matgin}
	\gamma_q\geq\tilde{\gamma}_0-\xi_q
\end{equation}
plays the same role as \eqref{eq:varian_seq_2}. Replacing \eqref{eq:varian_seq_2} with \eqref{eq:soft_matgin} and adding \eqref{eq:margin_constraint} into \eqref{opt:ldm}, we arrive at the following formulation
\begin{equation}
	\label{opt:odm_1}
	\begin{aligned}
		& &\underset{\boldsymbol{G},\ \boldsymbol{\xi},\ \boldsymbol{\varepsilon}}{\min}&\ \ \frac{1}{|\mathcal{Q}|}\sum_{q\in\mathcal{Q}}\xi_q+\nu\cdot\varepsilon_q\\
		& &s.t.&\ \ \gamma_q\geq\tilde{\gamma}_0-\xi_q,\ \gamma_q\leq\tilde{\gamma}_0+\varepsilon_q, \\
		& & &\ \ \xi_q\geq0,\ \varepsilon_q\geq0,\ \forall\ q\in\mathcal{Q}, \\
		& & &\ \ \boldsymbol{G}\succeq 0,\ \ \textit{rank}(\boldsymbol{G})\leq p.	
	\end{aligned}
\end{equation}
This optimization problem corresponds to dealing with such a loss function, $\forall\ q\in\mathcal{Q}$
\begin{equation}
	\label{eq:bisides-hinge}
	\begin{aligned}
			\ell_{\tilde{\gamma}_0,\nu}(\gamma_q)=\max(\tilde{\gamma}_0-\gamma_q,0)+\nu\cdot\max(\gamma_q-\tilde{\gamma}_0, 0).
	\end{aligned}
\end{equation}
The trading-off parameter $\nu$ in \eqref{opt:odm_1} can capture the asymmetry between the sign correctness and the dispersion of $\{\gamma_q\}_{q\in\mathcal{Q}}$. When $\nu=1$ and ignoring the semi-definite and rank constraints, \eqref{opt:odm_1} is similar to the support vector regression (\textit{SVR}) \cite{drucker1997support}. In \textit{SVR}, the $\tau$-insensitive loss
\begin{equation}
	\label{eq:insensitive-loss}
	|\zeta|_{\tau} = \left\{\begin{matrix}
	0, & \text{if }|\zeta|\leq \tau,\\ 
	|\zeta|-\tau, & \text{otherwise}  
\end{matrix}\right.
\end{equation}
produces two slack variables $\xi$ and $\xi^*$ for each training example to guard against outliers. As $\nu=1$, the loss function \eqref{eq:bisides-hinge} is explicitly the same loss function adopted by \textit{SVR} as $\tau=0$. In our formulation \eqref{opt:odm_1}, $\xi_q$ and $\varepsilon_q$ conduct the similar constraints like \textit{SVR} but have totally different meanings. 
All the training examples are used to learn the margin distribution in \eqref{opt:odm_1}, but the optimal solution of \textit{SVR} is only spanned by the support vectors which is sparse in the training data. Figure \ref{fig:distribution_loss} depicts the situations of the learned margin distribution graphically. Some theoretical results are provided at the end of this section.

\begin{theorem}
	\label{theom:finite_sample}
	Suppose that 
	$$
		\mathcal{G}_{\mu}=\{\boldsymbol{G}\in\mathbb{S}^n_+:\ \|\boldsymbol{G}\|_{\infty}\leq\mu, \|\boldsymbol{G}\|_*\leq \lambda\}
	$$
	and the true Gram matrix $\boldsymbol{G}^*\in\mathcal{G}_{\mu}$. Let $\hat{\boldsymbol{G}}$ be a solution of \eqref{opt:odm_1}. With probability at least $1-\delta$, it holds that
	\begin{equation}
		\begin{aligned}
		& &&\ \ R(\hat{\boldsymbol{G}})-R(\boldsymbol{G}^*)\\
		& &\leq&\ \ 4\mu(1+\nu)\sqrt{\frac{32np\log n}{|\mathcal{Q}|}}+8\mu(1+\nu)\sqrt{\frac{2\log\frac{2}{\delta}}{|\mathcal{Q}|}},
		\end{aligned}
	\end{equation}
	where $R(\cdot)$ is the risk, as for any $\boldsymbol{G}\in\mathcal{G}_\mu$
	$$
		R(\boldsymbol{G})=\mathbb{E}[\ell(\gamma_q)]=\frac{1}{|\mathcal{Q}|}\underset{q\in\mathcal{Q}}{\sum}p_q\ell(\gamma_q)+(1-p_q)\ell(-\gamma_q),
	$$
	$p_q=\mathbb{P}(y_q=1)$. Here the expectation respects to both the uniformly random selection of the quadruplet $q$ and its label $y_q$.
\end{theorem}

Theorem \ref{theom:finite_sample} says that $|\mathcal{Q}|$ must scale like $\Theta(pn\log n)$ which leads to the bounded error $R(\hat{\boldsymbol{G}})-R(\boldsymbol{G}^*)$. This result is consistent with known finite sample bounds \cite{jamieson2011low}. The details are provided in the supplementary materials. The generalization bound of margin distribution is a future direction.

\subsection{Optimization}
Consider that $\Delta_q$ is a linear operator on $\mathbb{S}^{n}_{+}$, $\Delta_q$ has its symmetric $n\times n$ matrix form in $\mathbb{S}^{n}_{+}$, the positive semi-definite cone of $n\times n $ symmetric matrix. Given a ordinal constraints $q=(i,j,l,k)$, $\boldsymbol{K}_q$ is the matrix form of $\Delta_q$ where 
\begin{equation}
	\boldsymbol{K}_q = \begin{blockarray}{crrrr}
	 &i&j&l&k\\
	\begin{block}{c(rrrr)}
	i&1&-1&0&0\\
	j&-1&1&0&0\\
	l&0&0&-1&1\\
	k&0&0&1&-1\\
	\end{block}
	\end{blockarray}
\end{equation}
and
\begin{equation}
	\Delta_q\boldsymbol{G}=\langle \boldsymbol{K}_q, \boldsymbol{G}\rangle=\text{tr}(\boldsymbol{K}_q\boldsymbol{G})=\textit{vec}(\boldsymbol{K}_q)^\top\textit{vec}(\boldsymbol{G}).
\end{equation}
With the trick that 
\begin{equation}
	\tilde{\gamma}_0-\gamma_q=y_qy_q\tilde{\gamma}_0-y_q\Delta_q\boldsymbol{G}=y_q(y_q\tilde{\gamma}_0-\Delta_q\boldsymbol{G}), 
\end{equation}we note
\begin{equation}
	\label{eq:auxiliary}
	\begin{aligned}
		& \boldsymbol{e}_{\mathcal{Q}} &=&\ \ [e_1,\dots,e_{|\mathcal{Q}|}]^\top\\
		& &=&\ \ \boldsymbol{y}_{\mathcal{Q}}\odot\boldsymbol{\Gamma}_0-
		\begin{pmatrix}
		\vdots\\
		\textit{vec}\left(\boldsymbol{K}_q\right)^\top\\
		\vdots
		\end{pmatrix}_{|\mathcal{Q}|\times n^2}\cdot\textit{vec}(\boldsymbol{G})\\
		& &\triangleq&\ \ \boldsymbol{y}_{\mathcal{Q}}\odot\boldsymbol{\Gamma}_0-\mathcal{K}\cdot\textit{vec}(\boldsymbol{G}),
	\end{aligned}
\end{equation}
where $\boldsymbol{y}_{\mathcal{Q}}\in\{-1,+1\}^{|\mathcal{Q}|}=[y_1,\dots,y_{|\mathcal{Q}|}]^\top$, $\boldsymbol{\Gamma}_0$ is a $|\mathcal{Q}|$-dimension vector with all entries are $\tilde{\gamma}_0$ and $\odot$ is the Hadamard product. Furthermore, we introduce the redundant variables to make the objective separable which can be solved by the ALM framework efficiently. The optimization is converted into:
\begin{equation}
	\label{opt:odm_3}
	\begin{aligned}
		& &\underset{\mathcal{T}}{\min}&\ \ \|(\boldsymbol{y}_{\mathcal{Q}}\odot \boldsymbol{e}_{1})_+\|^p_p+\|(\boldsymbol{y}_{\mathcal{Q}}\odot\boldsymbol{e}_{2})_+\|^p_p+\lambda\|\boldsymbol{G}_1\|_{*}\\
		& &s.t.&\ \ \boldsymbol{G} = \boldsymbol{G}_1,\ \boldsymbol{G}=\boldsymbol{G}_2,\ \boldsymbol{G}_2\succeq 0,\\
		& & &\ \  \boldsymbol{e}_{1} = \boldsymbol{e}_{\mathcal{Q}},\ \boldsymbol{e}_{2} = -\boldsymbol{e}_{\mathcal{Q}},
	\end{aligned}
\end{equation}
where $\mathcal{T}=\{\boldsymbol{G},\boldsymbol{G}_1, \boldsymbol{G}_2, \boldsymbol{e}_{1}, \boldsymbol{e}_{2}\}$ is the set of all the parameters to be solved and $\|\cdot\|_*$ is the nuclear norm which is the convex surrogate of matrix rank constraints. It is worth mentioning that (\ref{opt:odm_3}) is a convex optimization problem as the feasible set of each constraint is a convex set and the objective function is convex. The Lagrange function of (\ref{opt:odm_3}) can be written in the following form:
\begin{equation}
	\label{eq:lagrange_odm}
	\begin{aligned}
		& \mathcal{L}(\mathcal{T}) &=&\ \ \|(\boldsymbol{y}_{\mathcal{Q}}\odot \boldsymbol{e}_{1})_+\|+\|(\boldsymbol{y}_{\mathcal{Q}}\odot\boldsymbol{e}_{2})_+\|\\
		& &+&\ \ \Phi(\boldsymbol{z}_1, \boldsymbol{e}_1-\boldsymbol{e}_{\mathcal{Q}})+\Phi(\boldsymbol{z}_2, \boldsymbol{e}_2+\boldsymbol{e}_{\mathcal{Q}})\\
		& &+&\ \ \lambda\|\boldsymbol{G}_1\|_{*}\ +\ \Phi(\boldsymbol{Z}_3, \boldsymbol{G}-\boldsymbol{G}_1)\\
		& &+&\ \ \delta(\boldsymbol{G}_2\succeq 0)+\ \Phi(\boldsymbol{Z}_4, \boldsymbol{G}-\boldsymbol{G}_2),
	\end{aligned}
\end{equation}
with 
$$
\Phi(\boldsymbol{u},\boldsymbol{v})=\frac{\mu}{2}\|\boldsymbol{v}\|^2+\langle\boldsymbol{u},\boldsymbol{v}\rangle,\ \mu>0.
$$ 
$\|\cdot\|$ is $\ell_2$ norm for vector and the Frobenius norm for matrix. In addition, $\boldsymbol{z}_1, \boldsymbol{z}_2\in\mathbb{R}^{|\mathcal{Q}|}$ and $\boldsymbol{Z}_3, \boldsymbol{Z}_4\in\mathbb{R}^{n\times n}$ are Lagrange multipliers. $\delta$ is the Dirac delta function whose function value would be infinity if the condition is not satisfied. Below are the solutions to each sub-problem.
\\
\textbf{$\boldsymbol{e}_1$ sub-problem. }With the variables unrelated to $\boldsymbol{e}_1$ fixed, we have the sub-problem of $\boldsymbol{e}_1$:
\begin{equation}
	\label{opt:e1}
	\boldsymbol{e}^{(t+1)}_1 = \underset{\boldsymbol{e}_1}{\arg\min}\ \frac{1}{\mu^{(t)}}\|(\boldsymbol{y}_{\mathcal{Q}}\odot \boldsymbol{e}_{1})_+\|^p_p+\frac{1}{2}\|\boldsymbol{e}_1-\boldsymbol{s}^{(t)}_1\|^2_2,
\end{equation}
where 
$$
	\boldsymbol{s}^{(t)}_1=\boldsymbol{e}^{(t)}_{\mathcal{Q}}-\frac{\boldsymbol{z}^{(t)}_1}{\mu^{(t)}}.
$$
It's worth noting that $(\cdot)_+$ is a piece-wise linear function. Thus, to seek the minimum of each element in $\boldsymbol{e}_1$, we just need to pick the smaller value between $y_qe^1_{q}$ and $0$. 
The solution of (\ref{opt:e1}) is
\begin{equation}
	\label{eq:solution_e1_1}
	\boldsymbol{e}^{(t+1)}_1 = \Omega\left(\mathcal{S}_{\frac{1}{\mu^{(t)}}}[\boldsymbol{s}^{(t)}_1]\right) + \bar{\Omega}\left(\boldsymbol{s}^{(t)}_1\right),
\end{equation}
where $\Omega:\mathbb{R}^{\mathcal{Q}}\rightarrow\mathbb{R}^{\mathcal{Q}}$ is an indicator function as $[\Omega(\boldsymbol{w})]_q=\mathbb{I}(w_q>0)\cdot w_q,\ \boldsymbol{w}\in\mathbb{R}^{\mathcal{Q}}$ and $\bar{\Omega}$ is the complementary support of $\Omega$. The definition of shrinkage operator on scalars is $\mathcal{S}_{\tau>0}[u]=\textit{sign}(u)(|u|-\tau)_+$ and it is an  element-wise operator for vector and matrix.
\\
\textbf{$\boldsymbol{e}_2$ sub-problem. }Similarly, picking out the terms related to $\boldsymbol{e}_2$ gives the following sub-problem:
\begin{equation}
	\label{opt:e2}
	\boldsymbol{e}^{(t+1)}_2 = \underset{\boldsymbol{e}_2}{\arg\min}\ \frac{1}{\mu^{(t)}}\|(\boldsymbol{y}_{\mathcal{Q}}\odot \boldsymbol{e}_2)_+\|^p_p+\frac{1}{2}\|\boldsymbol{e}_2-\boldsymbol{s}^{(t)}_2\|^2_2,
\end{equation}
where
$$
	\boldsymbol{s}^{(t)}_2=-\boldsymbol{e}^{(t)}_{\mathcal{Q}}-\frac{\boldsymbol{z}^{(t)}_2}{\mu^{(t)}},
$$
and the solution of $\boldsymbol{e}_2$ sub-problem is just replaced the $\boldsymbol{s}^{(t)}_2$ with $\boldsymbol{s}^{(t)}_1$ in (\ref{eq:solution_e1_1}). 

\noindent\textbf{$\boldsymbol{G}$ sub-problem. }Dropping the terms independent on $\boldsymbol{G}$ leads to the following problem:
\begin{equation}
	\label{opt:G}
	\underset{\boldsymbol{G}}{\arg\min} \binom{\ \ \ \ \ \Phi(\boldsymbol{z}^{(t)}_1, \boldsymbol{e}^{(t)}_1-\boldsymbol{e}^{(t)}_{\mathcal{Q}})+\Phi(\boldsymbol{z}^{(t)}_2, \boldsymbol{e}^{(t)}_2+\boldsymbol{e}^{(t)}_{\mathcal{Q}})}{+\ \Phi(\boldsymbol{Z}^{(t)}_3, \boldsymbol{G}-\boldsymbol{G}^{(t)}_1)+\Phi(\boldsymbol{Z}^{(t)}_4, \boldsymbol{G}-\boldsymbol{G}^{(t)}_2)}.\\
\end{equation}
We have
\begin{equation}
	\begin{aligned}
		& & & 2(\mathcal{K}^\top\mathcal{K}+\boldsymbol{I})\cdot\textit{vec}(\boldsymbol{G})^{(t+1)}\\
		& &=&\ \ \mathcal{K}^\top\left(\boldsymbol{e}^{(t)}_1+\frac{1}{\mu^{(t)}}\boldsymbol{z}^{(t)}_1-\boldsymbol{e}^{(t)}_2-\frac{1}{\mu^{(t)}}\boldsymbol{z}^{(t)}_2\right)\\
		& &+&\ \ \textit{vec}\left(\boldsymbol{G}^{(t)}_1-\frac{1}{\mu^{(t)}}\boldsymbol{Z}^{(t)}_3+\boldsymbol{G}^{(t)}_2-\frac{1}{\mu^{(t)}}\boldsymbol{Z}^{(t)}_4\right),
	\end{aligned}
\end{equation}
and note the right-hand side as $\boldsymbol{w}$, we have $$\textit{vec}(\boldsymbol{G})^{(t+1)} = \frac{1}{2}(\mathcal{K}^\top\mathcal{K}+\boldsymbol{I})^{-1}\boldsymbol{w},$$ and $\boldsymbol{G}^{(t+1)}$ is the matrix form of $\textit{vec}(\boldsymbol{G})^{(t+1)}$.
\\
\textbf{$\boldsymbol{G}_1$ sub-problem. } There are two terms in (\ref{eq:lagrange_odm}) involving $\boldsymbol{G}_1$. The associated optimization problem of $\boldsymbol{G}_1$ is
\begin{equation}
	\label{opt:G1}
	\boldsymbol{G}_1^{(t+1)} = \underset{\boldsymbol{G}}{\arg\min}\ \lambda\|\boldsymbol{G}_1\|_{*}+\Phi(\boldsymbol{Z}_3, \boldsymbol{G}-\boldsymbol{G}_1),
\end{equation}
and solving this problem yields
\begin{equation}
	\label{eq:solution_G1}
	\boldsymbol{G}_1^{(t+1)} = \boldsymbol{U}\mathcal{S}_{\frac{\lambda}{\mu^{(t)}}}(\boldsymbol{\Sigma})\boldsymbol{V}^\top,
\end{equation}
where $$\boldsymbol{G}^{(t)}_1+\frac{1}{\mu^{(t)}}\boldsymbol{Z}^{(t)}_3=\boldsymbol{U\Sigma V}^\top$$ and $\mathcal{S}_\tau(\cdot)$ is the shrinkage operator.
\\
\textbf{$\boldsymbol{G}_2$ sub-problem. } Considering the potential asymmetric of $\boldsymbol{G}^{(t)}$, we claim that $\boldsymbol{G}_2$ is the nearest symmetric positive semi-definite matrix of $\boldsymbol{G}^{(t)}$ in Frobenius norm \cite{HIGHAM1988103}. By the following theorem, we show the explicit solution of $\boldsymbol{G}^{(t)}_2$.
\begin{theorem}
	Suppose that $\boldsymbol{A}\in\mathbb{R}^{n\times n}$, and let $\boldsymbol{B} = (\boldsymbol{A}+\boldsymbol{A}^\top)/2$, $\boldsymbol{C} = (\boldsymbol{A}-\boldsymbol{A}^\top)/2$ be the symmetric and skew-symmetric parts of $\boldsymbol{A}$ respectively. If we do polar decomposition of $\boldsymbol{B}$ as $\boldsymbol{B}=\boldsymbol{UH}$ where $\boldsymbol{U}$ is orthogonal matrix $\boldsymbol{UU}^\top=\boldsymbol{I}$ and $\boldsymbol{H}$ is positive semi-definite matrix, $\boldsymbol{X}_F = (\boldsymbol{B}+\boldsymbol{H})/2$ is the unique approximation of $\boldsymbol{A}$ in the Frobenius norm with positive semi-definite constraint, and the distance $\rho_F(\boldsymbol{A})$ in the Frobenius norm from $\boldsymbol{A}$ to $\mathbb{S}^n_+$ is
	\begin{equation*}
		\rho^2_F(\boldsymbol{A}) = \underset{\sigma_i(\boldsymbol{B})<0}{\sum}\sigma^2_i(\boldsymbol{B})+\|\boldsymbol{C}\|^2_F,
	\end{equation*}
	where $\sigma_i(\boldsymbol{B}),\ i=1,\dots,n$ is the eigenvalue of $\boldsymbol{B}$.
\end{theorem}
Consequently, the explicit solution of $\boldsymbol{G}_2^{(t)}$ is
\begin{equation}
	\label{eq:solution_G2}
	\begin{aligned}
		& \boldsymbol{G}_2^{(t)} &=&\ \ \frac{1}{4}\left(\boldsymbol{G}_{(t)}+\frac{1}{\mu}\boldsymbol{Z}_4+\boldsymbol{G}_{(t)}^\top+\frac{1}{\mu}\boldsymbol{Z}^\top_4\right)\\
		& &+&\ \ \frac{1}{4}\sqrt{\left(\boldsymbol{G}_{(t)}+\frac{1}{\mu}\boldsymbol{Z}_4\right)^\top\left(\boldsymbol{G}_{(t)}+\frac{1}{\mu}\boldsymbol{Z}_4\right)},
	\end{aligned}
\end{equation}
where $\sqrt{\boldsymbol{A}}$ is the square root of $\boldsymbol{A}\in\mathbb{S}^n_+$, $\boldsymbol{A} = \boldsymbol{VSV}^{-1}$ and $\sqrt{\boldsymbol{A}} = \boldsymbol{V}\boldsymbol{S}^{\frac{1}{2}}\boldsymbol{V}^{-1}$, $\boldsymbol{S}$ is a diagonal matrix and $\boldsymbol{S}^{\frac{1}{2}}$ is element-wise square root of $\boldsymbol{S}$.

\begin{algorithm}[ht]
	\caption{The ADMM method for solving $(\ref{opt:odm_3})$}
	\label{alg:admm_ome}
	\KwIn{$\mathcal{K}$, $\boldsymbol{y}_\mathcal{Q}$, $\gamma_0$, $\nu$, $\lambda$}
	\KwOut{$\boldsymbol{G}$}
	Initialize $\boldsymbol{G}_0$, $\mu$, $\boldsymbol{z}^{(0)}_1 = \boldsymbol{z}^{(0)}_2 = \boldsymbol{0}_{|\mathcal{Q}|}$, $t:=1$, $\boldsymbol{Z}^{(0)}_1=\boldsymbol{Z}^{(0)}_2=\boldsymbol{0}_{n\times n}$, calculate $\boldsymbol{e}_\mathcal{Q}$ via (\ref{eq:auxiliary})\;
	\While{Not Converged}
	{
		update $\boldsymbol{e}_1$, $\boldsymbol{e}_2$ by solving (\ref{opt:e1}) and (\ref{opt:e2})\;
		update $\boldsymbol{G}$ by solving (\ref{opt:G})\;
		update $\boldsymbol{G}_1$, $\boldsymbol{G}_2$ via (\ref{eq:solution_G1}) and (\ref{eq:solution_G2})\;
		calculate $\boldsymbol{e}_\mathcal{Q}$ via (\ref{eq:auxiliary})\;
		update multipliers and $\mu$ as
		\begin{equation}
			\begin{aligned}
				&\boldsymbol{z}^{(t+1)}_1 &=&\ \ \ \boldsymbol{z}^{(t)}_1+\mu^{(t)}(\boldsymbol{e}^{(t)}_1-\boldsymbol{e}_\mathcal{Q})\\
				&\boldsymbol{z}^{(t+1)}_2 &=&\ \ \ \boldsymbol{z}^{(t)}_2+\mu^{(t)}(\boldsymbol{e}^{(t)}_2+\boldsymbol{e}_\mathcal{Q})\\
				&\boldsymbol{Z}^{(t+1)}_1 &=&\ \ \ \boldsymbol{Z}^{(t)}_1+\mu^{(t)}(\boldsymbol{G}^{(t)}-\boldsymbol{G}^{(t)}_1)\\
				&\boldsymbol{Z}^{(t+1)}_2 &=&\ \ \ \boldsymbol{Z}^{(t)}_2+\mu^{(t)}(\boldsymbol{G}^{(t)}-\boldsymbol{G}^{(t)}_2)\\
				&\mu^{(t+1)} &=&\ \ \rho\mu^{(t)},\ \rho>1
			\end{aligned}
		\end{equation}
		$t:=t+1$\;
	}
\end{algorithm}

For clarity, the procedure of solving (\ref{opt:odm_3}) is outlined in Algorithm \ref{alg:admm_ome}. The algorithm would not be terminated until the change of objective value in two successive iterations is smaller than a threshold (in the experiments, $0.001$ is the default setting). 
{
	\begin{table*}[t]

		\caption{\label{tab:synthetic}Performance Comparison on synthetic dataset with $200$, $1000$ and $10000$ samples as training data, respectively}
		\centering
		\begin{subtable}[c]{0.8\columnwidth}
			\caption{$200$ samples}
			\centering
			\begin{tabular}{lcccc}
				\toprule
				algorithm & min    & median & max    & std   \\
				\midrule  
				GNMDS-$p$     & 0.419 & 0.447 & 0.476 & 0.016 \\
				STE-$p$       & 0.397 & 0.426 & 0.461 & 0.016 \\
				TSTE-$p$      & 0.440 & 0.468 & 0.498 & 0.014 \\
				DMOE & \textbf{0.372} & \textbf{0.390} & \textbf{0.410} & \textbf{0.011} \\ 
				\bottomrule
			\end {tabular}
			\label{tab:synthetic:200}
		\end{subtable}
		\begin{subtable}[c]{0.6\columnwidth}
			\caption{$1000$ samples}
			\centering
			\begin{tabular}{cccc}
				\toprule
				min    & median  & max    & std   \\
				\midrule  
				0.318 & 0.341 & 0.359 & 0.009 \\ 
				0.375 & 0.385 & 0.401 & \textbf{0.007} \\ 
				0.426 & 0.441 & 0.466 & 0.011 \\ 
				\textbf{0.281} & \textbf{0.298} & \textbf{0.305} & 0.008 \\ 
				\bottomrule
			\end {tabular}
			\label{tab1:synthetic:1000}
		\end{subtable}
		\begin{subtable}[c]{0.6\columnwidth}
			\caption{$10000$ samples}
			\centering
			\begin{tabular}{cccc}
				\toprule
				min    & median  & max    & std   \\
				\midrule  
				0.143 & 0.147 & 0.154 & \textbf{0.007} \\ 
				0.219 & 0.234 & 0.251 & \textbf{0.007} \\ 
				0.238 & 0.257 & 0.271 & 0.011 \\ 
				\textbf{0.142} & \textbf{0.146} & \textbf{0.151} & 0.008 \\ 
				\bottomrule
			\end {tabular}
			\label{tab1:synthetic:10000}
		\end{subtable}
	\end{table*}
}

{
	\begin{table*}[t]
		\caption{\label{tab:music}Performance Comparison on music artists dataset with $200, 500$, $1000$ and $5000$ samples as training data.}
		\centering
		\begin{subtable}[c]{0.8\columnwidth}
			\caption{$200$ samples}
			\centering
			\begin{tabular}{lcccc}
				\toprule
				algorithm & min    & median & max    & std   \\
				\midrule  
				GNMDS-$p$     & 0.391 & 0.403 & 0.416 & 0.007 \\
				STE-$p$       & 0.444 & 0.455 & 0.475 & 0.008 \\
				TSTE-$p$      & 0.416 & 0.436 & 0.458 & 0.011 \\
				DOME & \textbf{0.372} & \textbf{0.385} & \textbf{0.400} & \textbf{0.007} \\ 
				\bottomrule
			\end {tabular}
			\label{tab:music:200}
		\end{subtable}
		\begin{subtable}[c]{0.6\columnwidth}
			\caption{$1000$ samples}
			\centering
			\begin{tabular}{cccc}
				\toprule
				min    & median  & max    & std   \\
				\midrule  
				0.307 & 0.317 & 0.332 & 0.007 \\
				0.397 & 0.415 & 0.429 & 0.007 \\
				0.377 & 0.389 & 0.406 & 0.007 \\
				\textbf{0.281} & \textbf{0.291} & \textbf{0.307} & 0.007 \\ 
				\bottomrule
			\end {tabular}
			\label{tab1:music:1000}
		\end{subtable}
		\begin{subtable}[c]{0.6\columnwidth}
			\caption{$5000$ samples}
			\centering
			\begin{tabular}{lcccc}
				\toprule
				min    & median  & max    & std   \\
				\midrule  
				0.225 & 0.239 & 0.257 & \textbf{0.006} \\
				0.252 & 0.275 & 0.294 & 0.011 \\
				0.243 & 0.259 & 0.297 & 0.013 \\
				\textbf{0.216} & \textbf{0.227} & \textbf{0.244} & 0.007 \\ 
				\bottomrule
			\end {tabular}
			\label{tab1:music:5000}
		\end{subtable}
	\end{table*}
}

\section{Empirical Study}
In this section, we show the results of simulations and real-world data experiments to demonstrate the effectiveness of the proposed algorithms. As the existed margin-based ordinal embedding methods, such as GNMDS, STE and TSTE, just use triple-wise comparisons as the ordinal constraints, we treat triple-wise comparisons as the input of the proposed algorithm for fair competition. The triple-wise comparisons $\mathcal{T}=\{(i,j,k)\}$ is a special case of quadruplets which means $l=i$ in $q=(i,j,l,k)\in\mathcal{Q}$. The $\Delta_p$ of triplet $t=(i,j,k)$ is also a symmetric $n\times n$ matrix indicated by $(i,j,k)$ as 

\begin{equation}
	\boldsymbol{K}_t = \begin{blockarray}{crrr}
	 &i&j&k\\
	\begin{block}{c(rrr)}
	i&0&-1&1\\
	j&-1&1&0\\
	k&1&0&-1\\
	\end{block}
	\end{blockarray}\ \ .
\end{equation}

Replacing $\boldsymbol{K}_q$ with $\boldsymbol{K}_t$ in those sub-problems in Algorithm \ref{alg:admm_ome}, the proposed \textit{DMOE} method could handle the triplets set $\mathcal{T}$ as the ordinal constraints. The reproducible code can be found here\footnote{\url{https://github.com/alphaprime/DMOE}}.

\subsection{Simulation}
\noindent\textbf{Settings. }The synthesized dataset consists of $100$ points $\{\boldsymbol{x}_i\}_{i=1}^{100}\subset\mathbb{R}^{10}$, where $\boldsymbol{x}_i\sim\mathcal{N}(\boldsymbol{0}, \frac{1}{20}\boldsymbol{I})$, $\boldsymbol{I}\in \mathbb{R}^{10\times 10}$ is the identity matrix. The possible similarity triple comparisons are generated based on the Euclidean distances between $\{\boldsymbol{x}_i\}$. 
We randomly sample $|\mathcal{T}|=\{200, 500, 1,000, 10,000\}$ triplets as the training set and the test set is the rest of all triplets. The embedding dimension is fixed to $10$. 
\\
\noindent\textbf{Evaluation Metrics. }We employ the generalization error to evaluate generalization ability of various algorithms. As the learned Gram matrix $\boldsymbol{G}$ from partial triple comparisons set $\mathcal{T}\subset[n]^3$ may be generalized to unknown triplets, the percentage of held-out triplets which is not satisfied in the $\boldsymbol{G}$ is the generalization error of the learned embedding.
\\
\noindent\textbf{Competitors. }We compare the proposed algorithm with three well-known ordinal embedding methods: GNMDS \cite{agarwal2007generalized}, STE and TSTE \cite{vandermaaten2012stochastic}. Note that we adopt the optimization strategy proposed by \cite{DBLP:conf/nips/JainJN16}, which performs gradient descent with line search, and projects the Gram matrix onto the subspace spanned by the top $p$ eigenvalues at each step (i.e. setting the smallest $n-p$ eigenvalues to $0$). We call the three competitors: GNMDS-$p$, STE-$p$ and TSTE-$p$, correspondingly. The optimization problem of GNMDS is (\ref{opt:gnmds}). STE replaces the hinge loss by logistic loss in (\ref{opt:gnmds}) and adopts Gaussian kernel to predict the label:
\begin{equation*}
	\label{eq:logistic}
	l_{\text{ste}}(\boldsymbol{G}^{(t)}, y_p) = \log(1+\exp(\gamma^{(t)}_p)),
\end{equation*} 
where $\gamma^{(t)}_p=\Delta_p\boldsymbol{G}^{(t)}$. TSTE employs the heavy-tailed Student-t kernel:
\begin{equation*}
	\label{eq:student}
	l_{\text{tste}}(\boldsymbol{G}^{(t)}, y_p) = \log\left(1+\left(\gamma^{(t)}_p\right)^{-\frac{\alpha+1}{2}}\right).
\end{equation*}
The regularization parameters of the competitors are tuned for the best performance under the different settings. 

\begin{figure}[thb!]
	\centering
	\begin{subfigure}[b]{.3\columnwidth}
		\includegraphics[width = \textwidth]{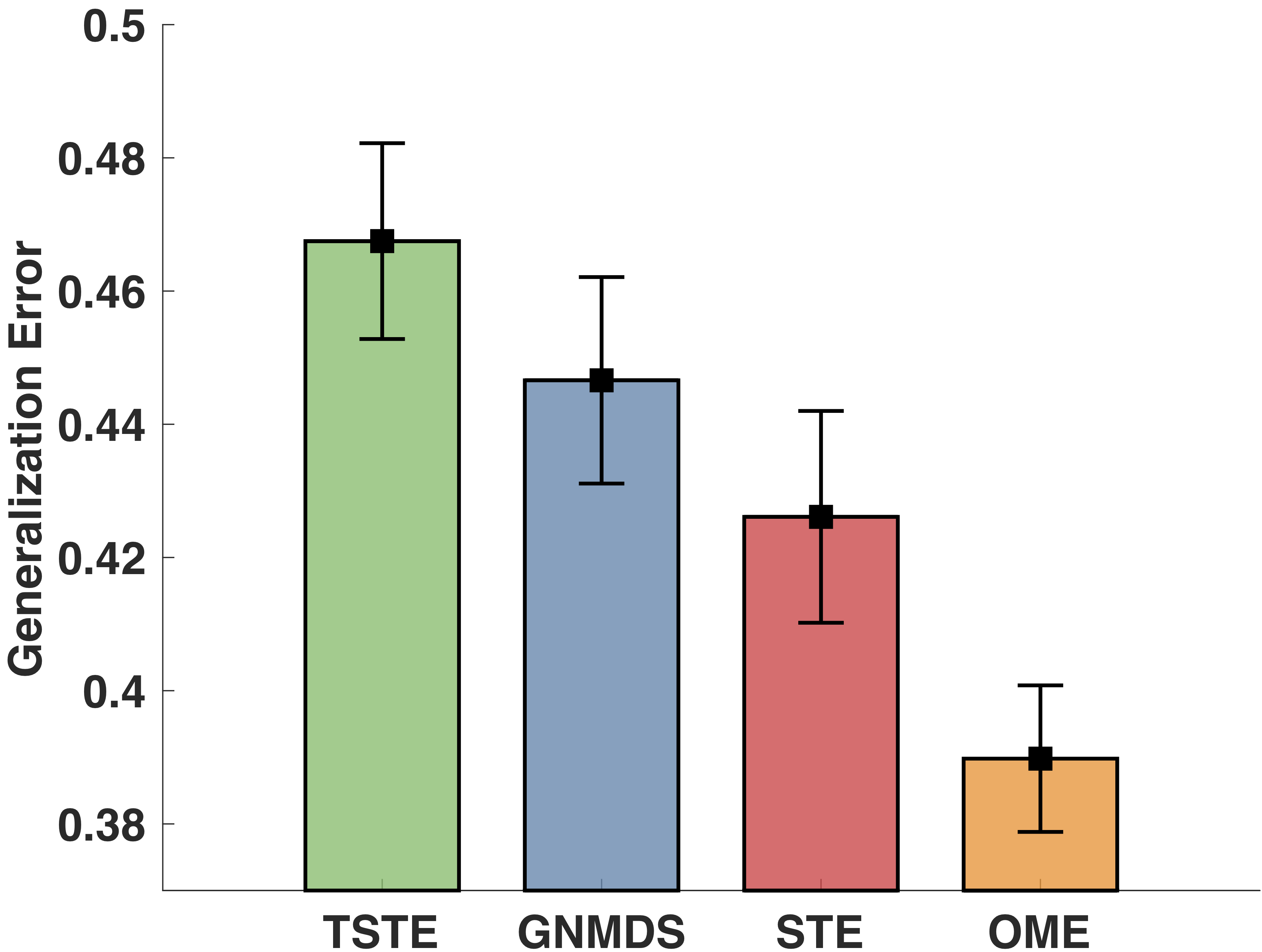}
		\caption{$200$}
		\label{fig:syth:200:error}
	\end{subfigure}
	\begin{subfigure}[b]{.3\columnwidth}
		\includegraphics[width = \textwidth]{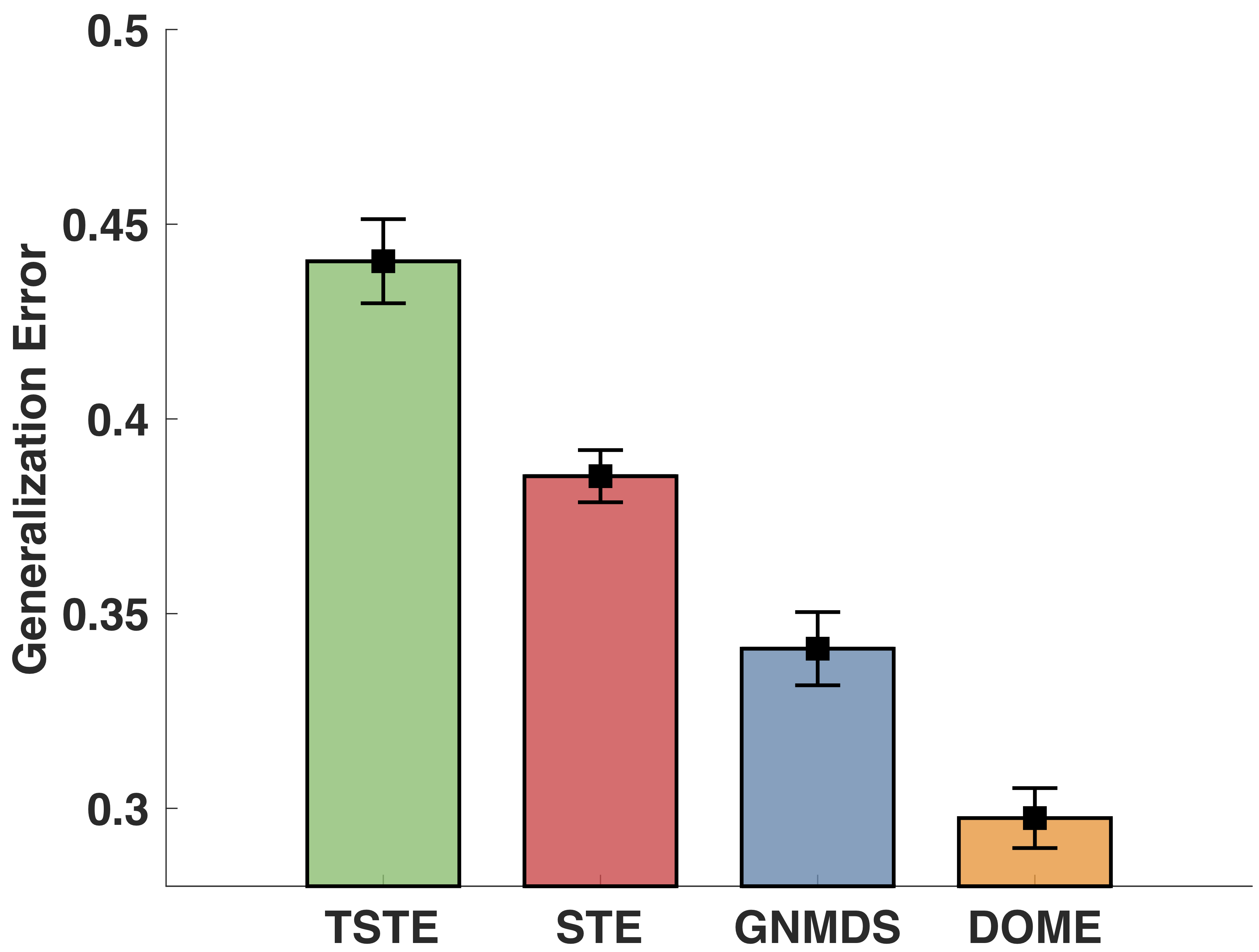}
		\caption{$1000$}
		\label{fig:syth:100:error}
	\end{subfigure}
	\begin{subfigure}[b]{.3\columnwidth}
		\includegraphics[width = \textwidth]{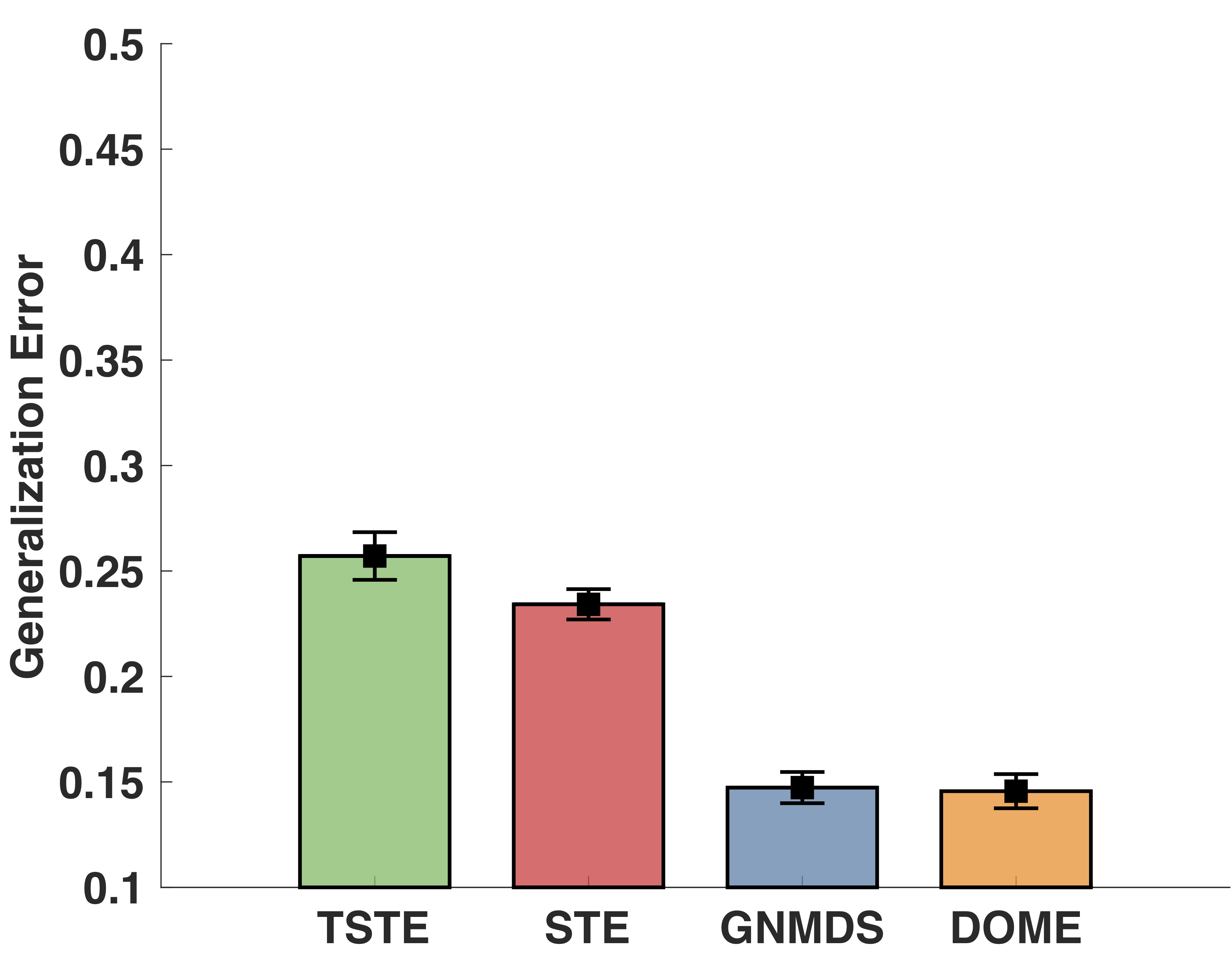}
		\caption{$10000$}
		\label{fig:syth:1000:error}
	\end{subfigure}
	\caption{Generalization errors of DMOE, GNMDS-$p$, STE-$p$ and TSTE-$p$ on the synthetic dataset.}
	\label{fig:synthetic}
\end{figure}
\noindent\textbf{Results. }From Figure \ref{fig:synthetic} and Table \ref{tab:synthetic}, the following phenomena can be observed. First of all, the generalization ability of all methods would be improved when the number of training samples increases. The decrease of standard derivation also improves the stability. Moreover, the proposed algorithm shows better generalization performance than the traditional methods in all four settings. Compared with GNMDS-$p$/STE-$p$/TSTE-$p$ which need more training samples, our method can achieve better results with fewer training samples. This is our main motivation to optimize the margin distribution instead of maximizing the minimum margin like the classic methods. Third, the results of GNMDS-$p$ verifies that only maximizing the minimum margin would not necessarily lead to better generalization performances as the STE-$p$ is better than GNMDS when train samples are few.

\begin{figure}[thb!]
	\centering
	\begin{subfigure}[b]{.3\columnwidth}
		\includegraphics[width = \textwidth]{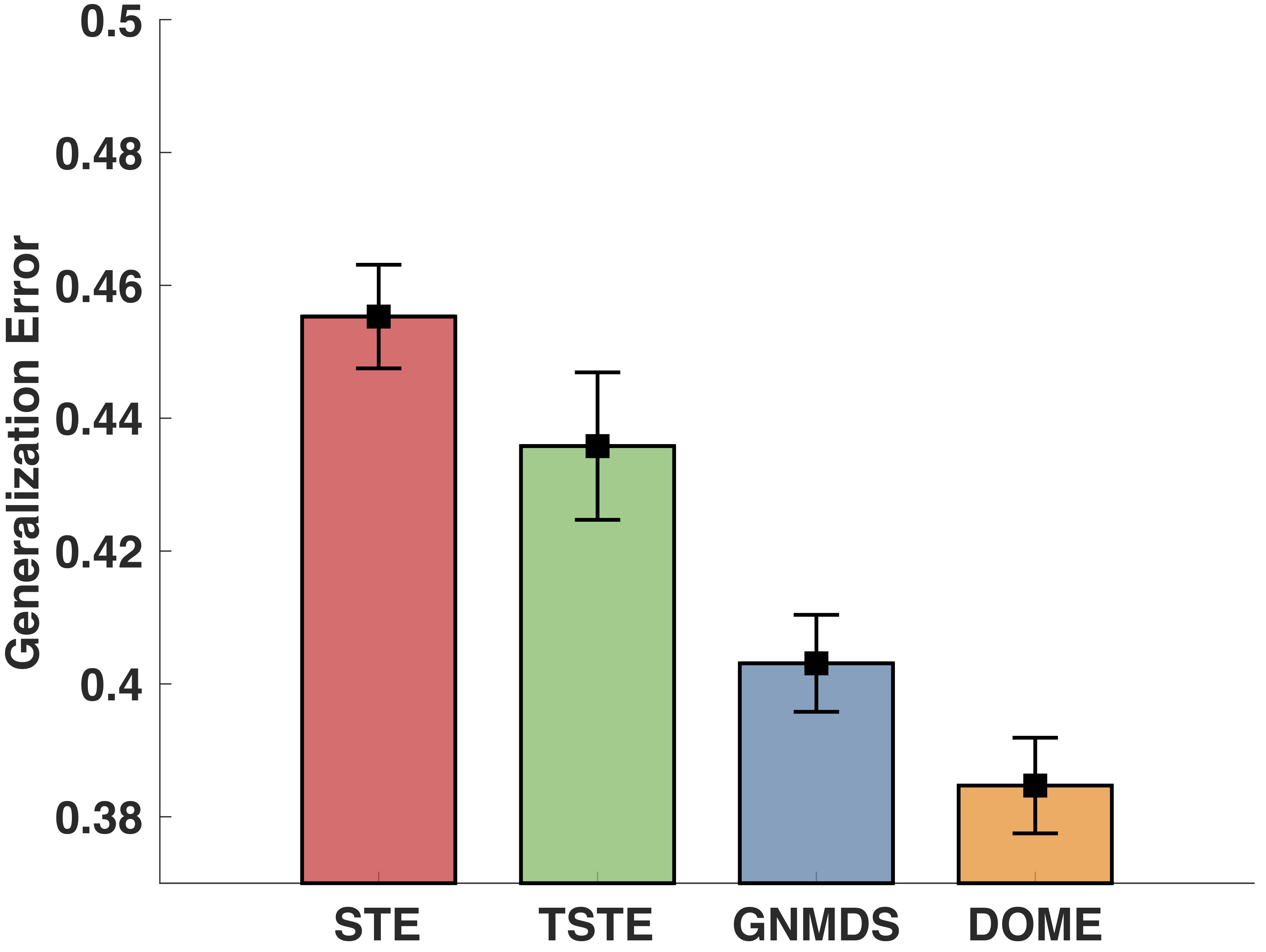}
		\caption{$200$}
		\label{fig:music:200:error}
	\end{subfigure}
	\begin{subfigure}[b]{.3\columnwidth}
		\includegraphics[width = \textwidth]{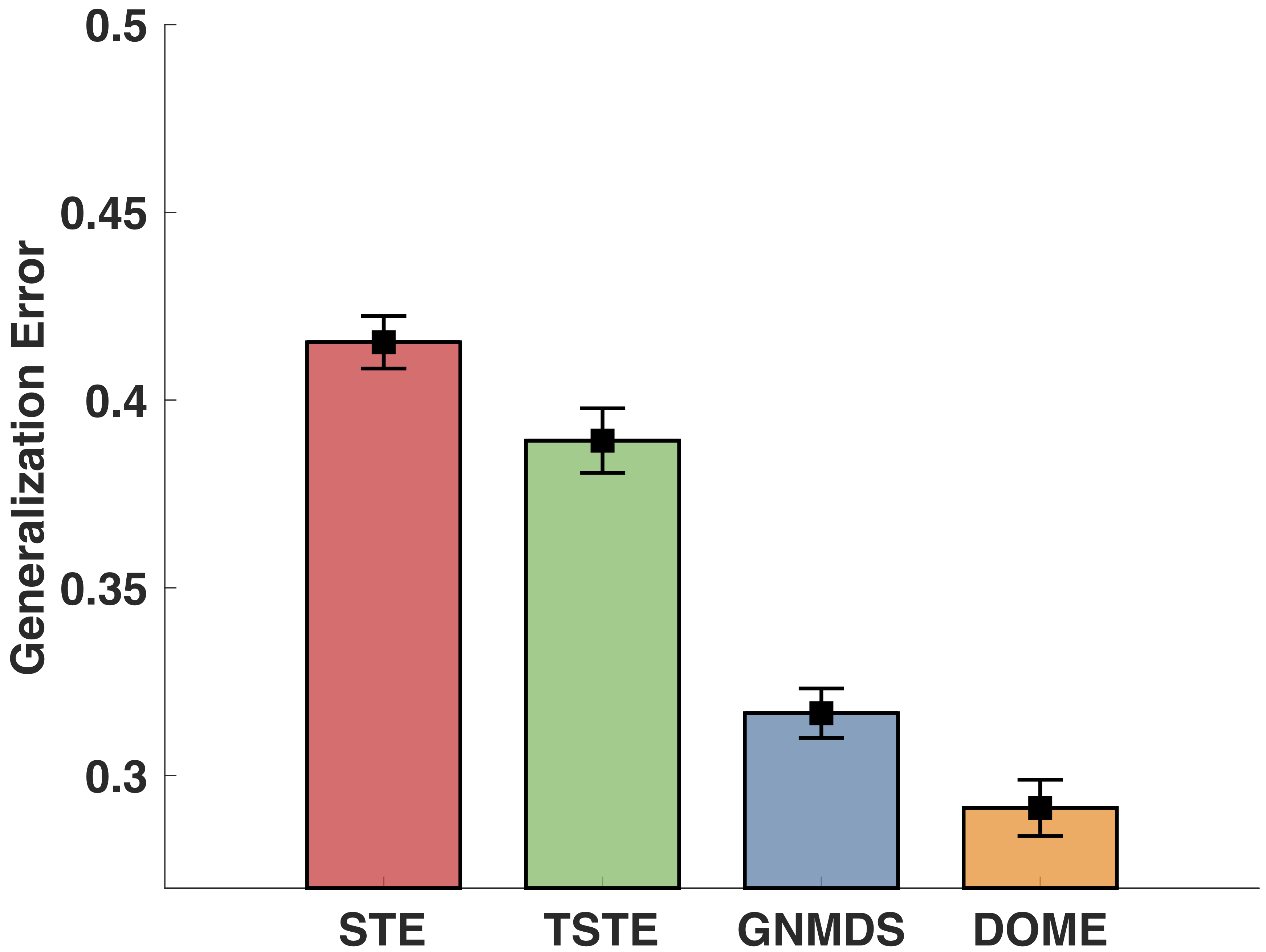}
		\caption{$1000$}
		\label{fig:music:100:error}
	\end{subfigure}
	\begin{subfigure}[b]{.3\columnwidth}
		\includegraphics[width = \textwidth]{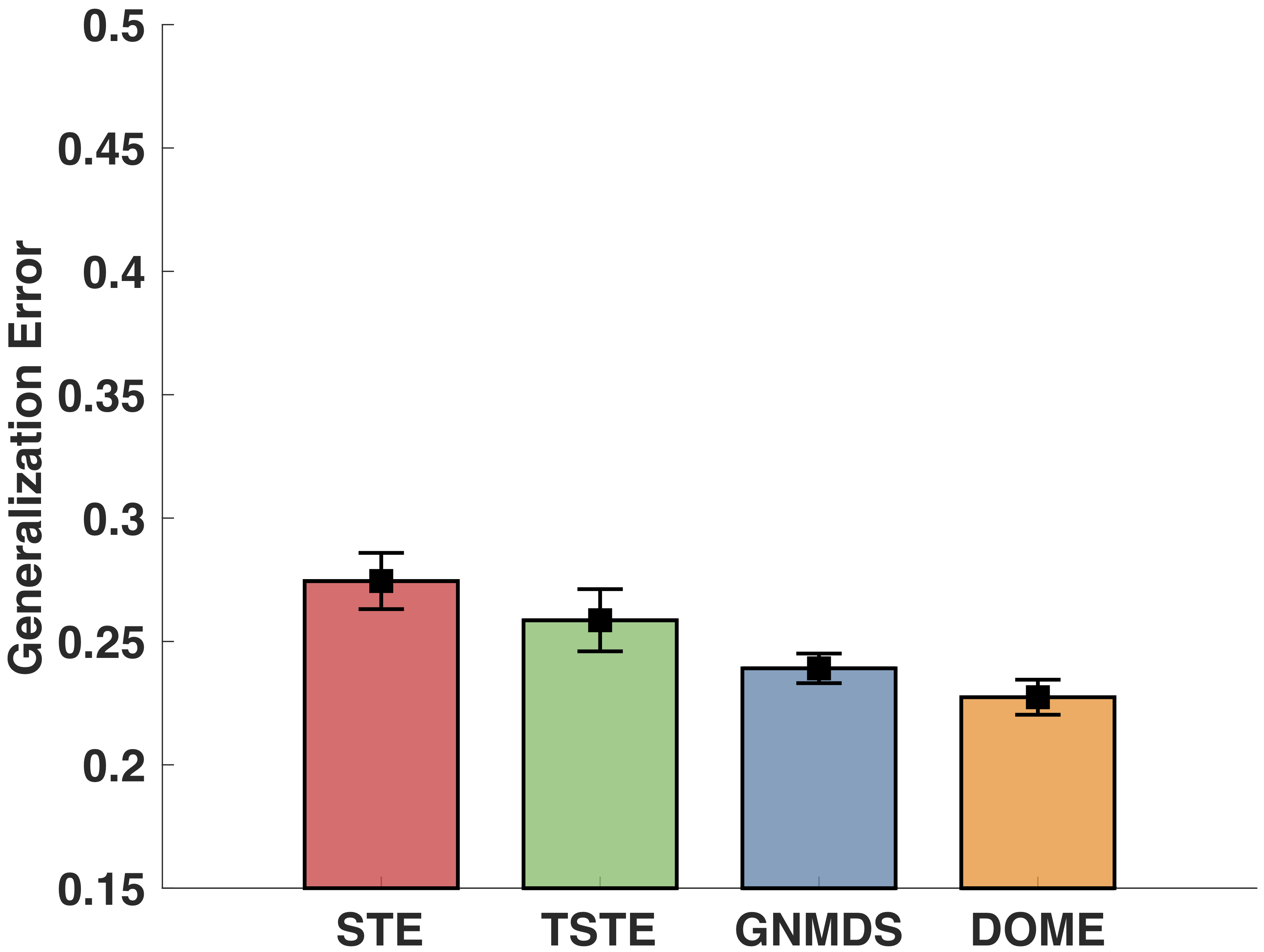}
		\caption{$5000$}
		\label{fig:music:5000:error}
	\end{subfigure}
	\caption{Generalization errors of DMOE, GNMDS, STE and TSTE on the music artists dataset.}
	\label{fig:music}
\end{figure}

\subsection{Music Artist Data}

\noindent\textbf{Settings. }The music artist data is collected by \cite{ellis2002quest} via a web-based survey in which $1,032$ users provided $213,472$ triplets on the similarity of $412$ music artists. We use the data pre-processed by \cite{vandermaaten2012stochastic} which includes only $9,107$ triplets for $n=400$ artists. The size of training samples is variant from $200$ to $5,000$ and the rest of triplets are treated as test set. The desired dimension of embedding is $d=9$ as these music artists can be classified by genre into $9$ categories. 

\noindent\textbf{Results. }According to the experimental results, Figure \ref{fig:music} and Table \ref{tab:music}, we have the following observations. \textit{DMOE} still shows better prediction result than GNMDS-$p$/STE-$p$/TSTE-$p$ with the same number of noisy training samples. To achieve the same generalization error, \textit{DMOE} needs the smallest number of training samples and STE-$p$/TSTE-$p$ need five times more than \textit{DMOE}. This real-world data experiment verifies the proposed method, \textit{DMOE}, has strong generalization for ordinal embedding with small training samples. Although this dataset contains noise triplets and it is well-known that the calculation of mean and the variance is sensitive, the proposed method show the same magnitude of standard deviation and its results are not damaged by the potential wrong training samples. The robustness is still an open problem in ordinal embedding, and this is our future work. 

\section{Conclusion}
The classical ordinal embedding algorithms always need a large number of labeled data to predict unknown similarity relationship among items from learned embedded points. As collecting high-quality, large-scale labeled data from human is a hard task, generalization ability is the main challenge when we could only access small numbers of relative comparisons. Incorporating margin distribution learning paradigm gives birth to a novel algorithm for ordinal embedding, namely \textit{DMOE}. Comprehensive experiments on synthetic dataset and real-world dataset validate the superiority of our method to traditional methods which need more training data to achieve the same generalization.

\section*{Acknowledgment}
The research of Ke Ma and Xiaochun Cao is supported by the National Key R\&D Program of China (Grant No. 2016YFB0800603), the Key Program of the Chinese Academy of Sciences (No. QYZDB-SSW-JSC003) and the National Natural Science Foundation of China (No.U1636214, U1605252, 61733007). The research of Qianqian Xu is supported in part by the National Natural Science Foundation of China (No.61672514, 61390514, 61572042), the Beijing Natural Science Foundation (4182079), the Youth Innovation Promotion Association CAS, and the CCF-Tencent Open Research Fund.

\bibliographystyle{aaai}
\bibliography{aaai19}

\end{document}